\documentclass{article}

\usepackage{microtype}
\usepackage{graphicx}
\usepackage{subfigure}
\usepackage{booktabs} % for professional tables
\usepackage{mathtools}
\usepackage{xfrac}

\usepackage{hyperref}

\usepackage[accepted]{icml2019}

\icmltitlerunning{How Can We Be So Dense?}

\hypersetup{draft}
\begin{document}

\twocolumn[
\icmltitle{How Can We Be So Dense? The Benefits of Using Highly Sparse Representations}



\icmlsetsymbol{equal}{*}

\begin{icmlauthorlist}
\icmlauthor{Subutai Ahmad}{nu}
\icmlauthor{Luiz Scheinkman}{nu} \\
Numenta, Redwood City, California, USA
\end{icmlauthorlist}

\icmlaffiliation{nu}{ }

\icmlcorrespondingauthor{Subutai Ahmad, Luiz Scheinkman}{[sahmad, lscheinkman]@numenta.com}

\icmlkeywords{Machine Learning, ICML, Sparse Representations}

\vskip 0.3in
]



\printAffiliationsAndNotice{ }  % leave blank if no need to mention equal contribution

\begin{abstract}
Most artificial networks today rely on dense representations, whereas biological networks rely on sparse representations. In this paper we show how sparse representations can be more robust to noise and interference, as long as the underlying dimensionality is sufficiently high. A key intuition that we develop is that the ratio of the operable volume around a sparse vector divided by the volume of the representational space decreases exponentially with dimensionality. We then analyze computationally efficient sparse networks containing both sparse weights and activations. Simulations on MNIST and the Google Speech Command Dataset show that such networks demonstrate significantly improved robustness and stability compared to dense networks, while maintaining competitive accuracy. We discuss the potential benefits of sparsity on accuracy, noise robustness, hyperparameter tuning, learning speed, computational efficiency, and power requirements.
\end{abstract}

\section{Introduction}
\label{introduction}

The literature on sparse representations in neural networks dates back many decades, with neuroscience as one of the primary motivations. In 1988 Kanerva proposed the use of sparse distributed memories \cite{Kanerva1988} to model the highly sparse representations seen in the brain. In 1997, \cite{Olshausen1997} showed that incorporating sparse priors and sparse cost functions in encoders can lead to receptive field representations that are remarkably close to what is observed in the primate visual cortex. More recently \cite{Lee2008,Chen2018b} showed hierarchical sparse representations that qualitatively lead to natural looking hierarchical feature detectors. \cite{Lee2009,Nair2009,NIPS2013_5059,Rawlinson2018} showed that introducing sparsity terms can sometimes lead to improved test set accuracies. 

Despite the above literature the majority of neural networks today rely on dense representations. One exception is the pervasive use of dropout \cite{Srivastava2014} as a regularizer. Dropout randomly ``kills'' a percentage of the units (in practice usually $50\%$) on every training input presentation. Variational dropout techniques tune the dropout rates individually per weight \cite{Molchanov2017}. Dropout introduces random sparse representations during learning, and has been shown to be an effective regularizer in many contexts. 

In this paper we discuss certain inherent benefits of high dimensional sparse representations. We focus on robustness and sensitivity to interference. These are central issues with today's neural network systems where even small \cite{Szegedy2013} and large \cite{Rosenfeld2018} perturbations can cause dramatic changes to a network's output. We offer two main contributions. First, we analyze high dimensional sparse representations, and show that such representations are naturally more robust to noise and interference from random inputs. When matching sparse patterns, corrupted versions of a pattern are ``close'' to the original whereas random patterns are exponentially hard to match.

Our second contribution is an efficient construction of sparse deep networks that is designed to exploit the above properties. We implement networks where the weights for each unit in a layer randomly sample from a sparse subset of the source layer below. In addition the output of each layer is constrained such that only the $k$ most active units are allowed to be non-zero, where $k$ is much smaller than the number of units in that layer. In these networks, the number of non-zero products for each layer is approximately $(\text{sparsity of layer } i) \times (\text{sparse weights of layer } i+1)$.  This formulation results in simple differentiable sparse layers that can be dropped into both standard linear and convolutional layers.

We demonstrate significantly improved robustness to noise for MNIST and the Google Speech Commands dataset, while maintaining competitive accuracy in the standard zero noise scenario. We discuss the number of weights used by sparse networks in these datasets, and the impact of additional pruning. Our work extends the existing literature on sparse networks and pruning (see Section~\ref{discussion} for a comparison with some prior work). At the end of the paper we discuss some possible areas for future work.

\section{High Dimensional Sparse Representations}
\label{HighDim}

In this section we develop some basic properties of sparse representations as they relate to noise robustness and interference. In a typical neural network an input vector is matched against a stored weight vector using a dot product. This is then followed by a threshold-like non-linearity such as $\tanh(\cdot)$ or $\text{ReLU}(\cdot)$. 

Ideally we would like the outputs of each layer to be invariant to noise or corrupted inputs.  When comparing two sparse vectors via a dot product, the results are unaffected by the zero components of either vector. A key quantity we consider is the ratio of the matching volume around a prototype vector divided by the volume of the whole space. The larger the match volume around a vector, the more robust it is to noise. The smaller the ratio, the less likely it is that random inputs can affect the match.

\subsection{Matching Sparse Binary Vectors}

We quantify the above ratio using binary vectors (following our previous work in \cite{Ahmad2016}). In this section we show that the ratio decreases exponentially with increased dimensionality, while maintaining a large match volume.  Let $\boldsymbol{x}$ be a binary vector of length $n$, and let $|\boldsymbol{x}|$ denote the number of non-zero entries. The dot product $\boldsymbol{x}_i \cdot \boldsymbol{x}_j$ counts the overlap, or number of shared bits, between two such vectors. We would like to understand the probability of two vectors having significant overlap, i.e. overlap greater than some threshold $\theta$.

We define the overlap set, $\Omega^n(\boldsymbol{x}_i, b, k)$, as the set of all vectors of size $k$ that have exactly $b$ bits of overlap with $\boldsymbol{x_i}$. The number of such vectors can be calculated as:

\begin{equation}
\label{omega}
|\Omega^n(\boldsymbol{x}_i, b, k)| = 
{\binom{|\boldsymbol{x}_i|}{b}} 
{\binom{n-|\boldsymbol{x}_i|}{k - b}}
\end{equation}

The left half of the above product counts all the ways we can select exactly $b$ bits out of active bits in $|\boldsymbol{x}_i|$. The right half counts the number of ways we can select the remaining $k-b$ bits from the components of $\boldsymbol{x}_i$ that are zero. The product of these two quantities represents the number of all vectors with exactly $b$ bits of overlap with $|\boldsymbol{x}_i|$. We can now count the number of vectors that match $\boldsymbol{x}_i$, i.e. where $\boldsymbol{x}_i \cdot \boldsymbol{x}_j \geq \theta$ as: 

\begin{equation}
\sum_{b=\theta}^{|\boldsymbol{x}_i|} \mid \Omega^n(\boldsymbol{x}_i,b,|\boldsymbol{x}_j|)\mid
\end{equation}

If we select vectors from a uniform random distribution, the probability of significant overlap can be calculated as:

\begin{equation}
P(\boldsymbol{x}_i \cdot \boldsymbol{x}_j \geq \theta) = 
\frac{\sum_{b=\theta}^{|\boldsymbol{x}_i|} \mid \Omega^n(\boldsymbol{x}_i,b,|\boldsymbol{x}_j|)\mid}
{\binom{n}{|\boldsymbol{x}_j|}}
\label{p1}
\end{equation}

where $\binom{n}{|\boldsymbol{x}_j|}$ is the set of all possible comparison vectors.  

\begin{figure}[t]
\vskip 0.2in
\begin{center}
\centerline{\includegraphics[width=\columnwidth]{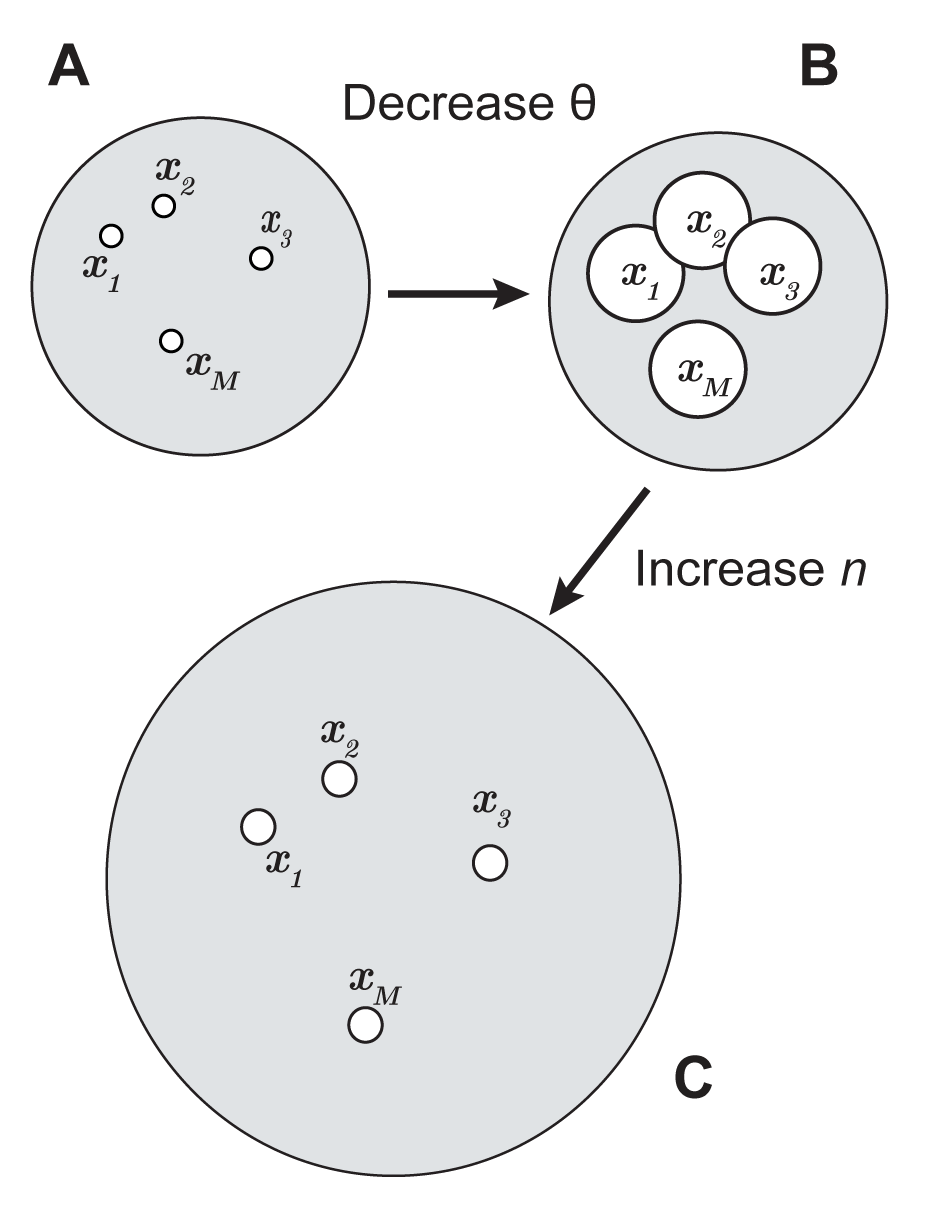}}
\caption{An illustration of the conceptual effect of decreasing the match threshold $\theta$ and increasing $n$, the dimensionality. The large grey circles denote the universe of possible patterns. The smaller circles each represent the set of matches around one vector. When $\theta$ is high (\textbf{A}), very few random vectors can match these vectors (small white circles). As you decrease $\theta$, the set of potential matches increases (larger white circles in \textbf{B}). If you then increase $n$, the universe of possible patterns increases, and the relative sizes of the white circles shrink rapidly.}
\label{circles}
\end{center}
\vskip -0.2in
\end{figure}

\subsection{Impact of Dimensionality and Sparsity}

Two key factors in Eq.~\ref{p1} are the number of non-zero components, $|\boldsymbol{x}_i|$, and the dimensionality, $n$. Figure~\ref{circles} provides an intuitive description of their impact. Assume we have $M$ prototype vectors, and we want to match noisy versions of these vectors. Around each prototype there is a set of matching vectors. If the threshold is very high, the set of matching vectors is small (illustrated by the small circles in Figure~\ref{circles}A) and there will be quite a bit of space between these sets. As you decrease $\theta$ matching is less strict and you can match noisier versions of each prototype. The cost is that the chance of matching the other vectors also increases because there is less free space in between (Figure~\ref{circles}B). It turns out that for sparse vectors, this cost is offset as you increase $n$. That is, as $n$ increases, the denominator in Eq.~\ref{p1} (and the corresponding "free" space) increases much faster than the numerator. For a fixed sparsity level, you can maintain highly tolerant matches without the cost of additional false positives simply by increasing the dimensionality.

\begin{figure}[ht]
\vskip 0.2in
\begin{center}
\centerline{\includegraphics[width=\columnwidth]{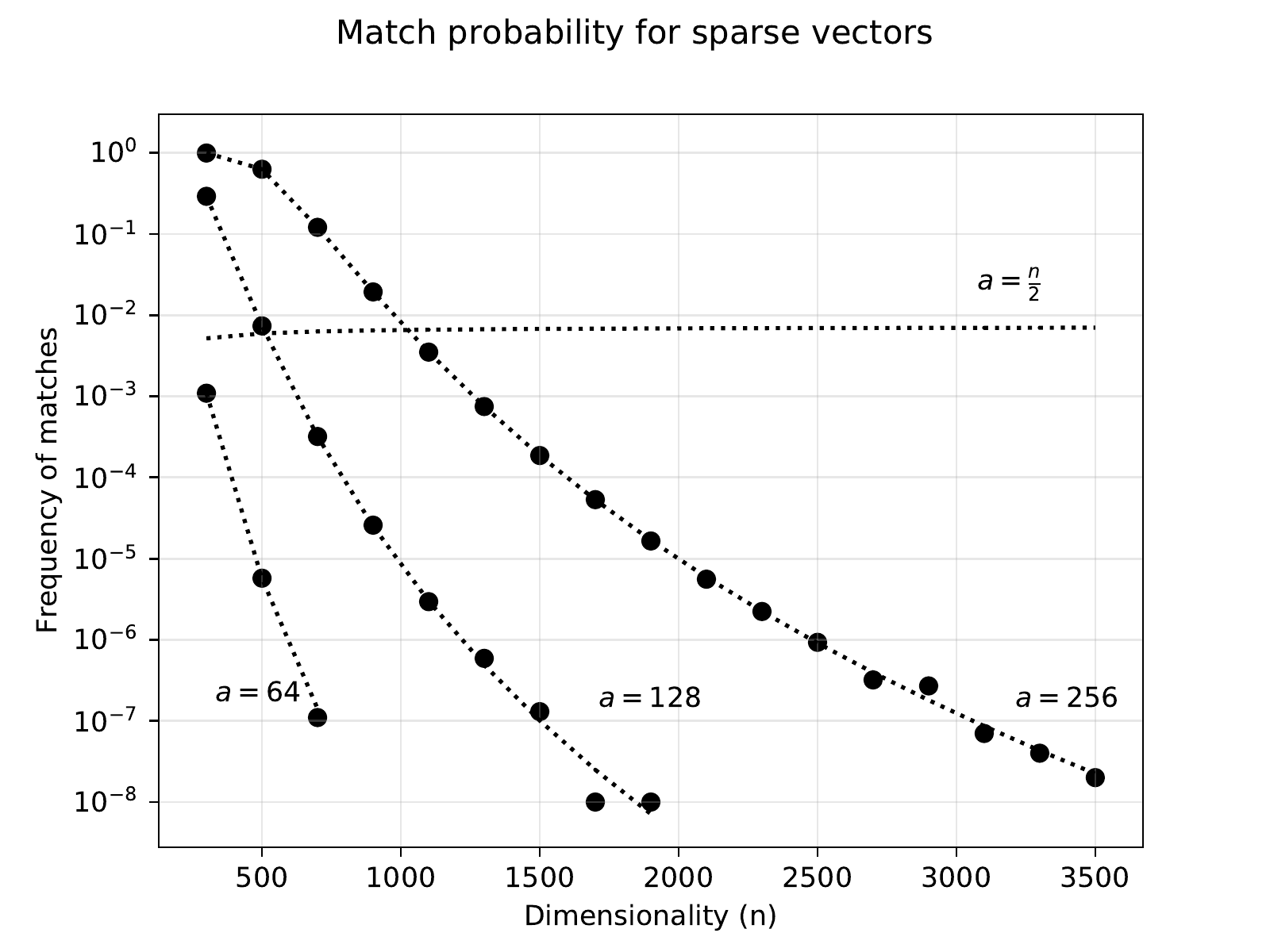}}
\caption{The probability of matches to random binary vectors (with $a$ active bits) as a function of dimensionality, for various levels of sparsity. The probability decreases exponentially with $n$. Black circles denote the observed frequency of a match (based on a large number of trials). The dotted lines denote the theoretically predicted probabilities using Eq.~\ref{p1}.}
\label{sdr}
\end{center}
\vskip -0.2in
\end{figure}

Fig~\ref{sdr} illustrates this trend for some example sparsities. In this figure we simulated matching with random vectors and plotted match rates with random vectors as a function of the number of active bits and the underlying dimensionality. In the simulation we repeatedly generated a random prototype vector with $|\boldsymbol{x}_i|=24$ bits on and then attempted to match against random test vectors with $a$ bits on. We matched using a threshold $\theta$ of 12 which meant that even vectors that were up to 50\% different from $\boldsymbol{x}_i$ would match. We varied $a$ and the dimensionality of the vectors, $n$.

The chart shows that for sparse binary vectors, match rates with random vectors drop rapidly as the underlying dimensionality increases. The horizontal line indicates the probability of matching $\boldsymbol{x}_i$ against dense vectors, with $a=n/2$. The probability of dense matches stays relatively high and unaffected by dimensionality, indicating that both sparseness and high dimensionality are key to robust matches. In \cite{Ahmad2016} we develop additional properties, including the probability of false negatives.

\subsection{Matching Sparse Scalar Vectors}

\begin{figure*}
\vskip 0.2in
\begin{center}
\centerline{\includegraphics[width=\textwidth]{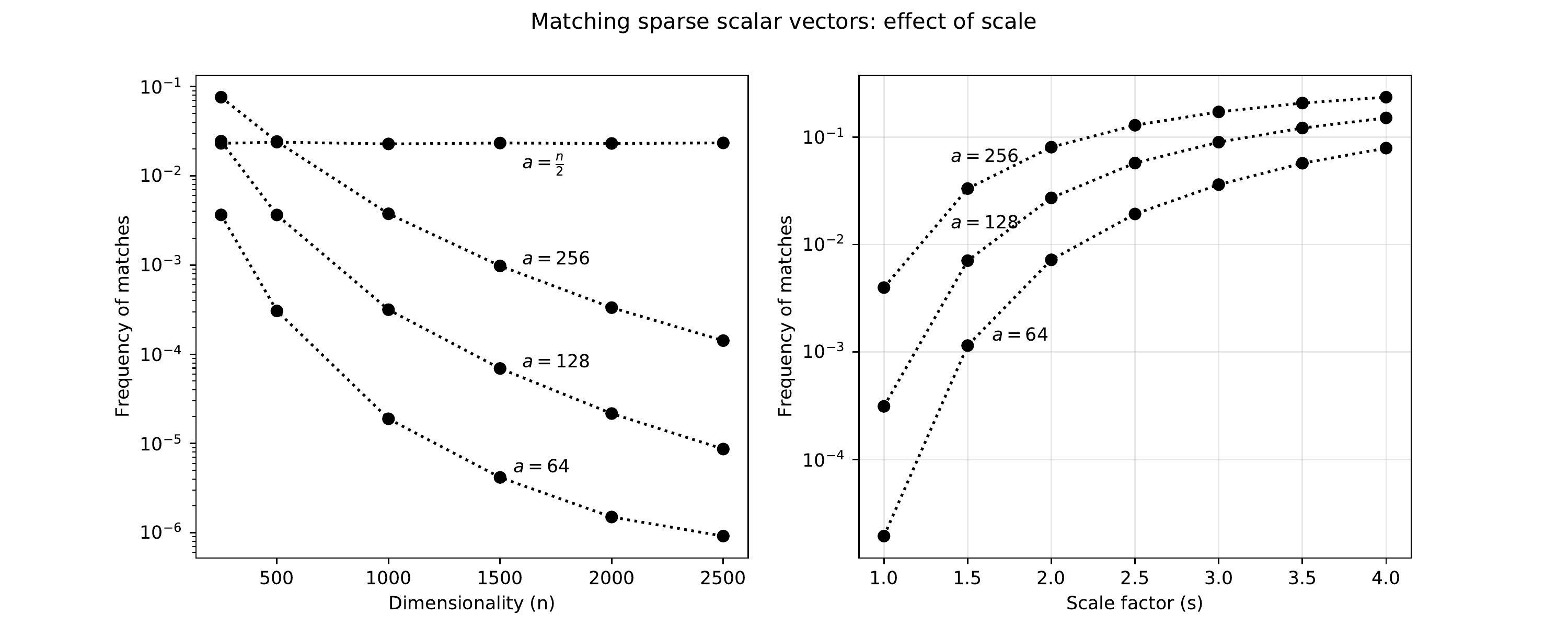}}
\vskip -0.2in
\caption{\textbf{Left}: The probability of matches to random scalar vectors (with $a$ non-zero components) as a function of dimensionality, for various levels of sparsity. The probability of false matches decreases exponentially with $n$. Note that the probability for a dense vector, $a=\frac{n}{2}$ stays relatively high, and does not decrease with dimensionality. \textbf{Right}: The impact of scale on vector matches with a fixed $n=1000$. The larger the scaling discrepancy, the higher the probability of a false match.}
\label{scalar-sdr}
\end{center}
%\vskip -0.2in
\end{figure*}

Deep networks operate on scalar vectors, and in this section we consider how the above ideas apply to sparse scalar representations.  Binary and scalar vectors are similar in that the components containing zero do not affect the dot product, and thus the combinatorics in Eq.~\ref{p1} are still applicable. Eq.~\ref{omega} represents the set of scalar vectors where the number of non-zero multiplies in the dot product is exactly $b$, and Eq.~\ref{p1}  represents the probability that the number of non-zero multiplies is $>=\theta$. However, an additional factor is the distribution of scalar values. If components in one vector are extremely large relative to $\theta$, the likelihood of a significant match will be high even with a single shared non-zero component.

We wanted to see if the exponential drop in random matches for binary vectors, demonstrated by Figure~\ref{sdr}, can be obtained using scalar vectors, and if so, the conditions under which they hold.  Let $\boldsymbol{x}_w$ and $\boldsymbol{x}_i$ represent two sparse vectors such that $\| \boldsymbol{x}_w \|_0$ and $\| \boldsymbol{x}_i \|_0$ counts the number of non-zero entries in each.  Let each non-zero component be independent and sampled from the distributions $P_{\theta_w}(x{_w})$ and $P_{\theta_i}(x{_i})$. The probability of a significant match is then:

\begin{multline}
P(\boldsymbol{x}_w \cdot \boldsymbol{x}_i \geq \theta) = \\
\frac{\sum_{b=\theta}^{\|\boldsymbol{x}_w\|_0} p_b \mid \Omega^n(\boldsymbol{x}_w,b,\| \boldsymbol{x}_i \|_0)\mid }
{\binom{n}{\|\boldsymbol{x}_i \|_0}}
\label{p1scalar}
\end{multline}

where $p_b$ is the probability that the dot product is $>= \theta$ given that the overlap is exactly $b$ components:

\begin{equation}
p_b = P(\boldsymbol{x}_w \cdot \boldsymbol{x}_i \geq \theta \mid \|\boldsymbol{x}_w \cdot \boldsymbol{x}_i\|_0 = b)
\end{equation}

There does not appear to be a closed form way to compute $p_b$ for normal or uniform distributions so we resort to simulations that mimic our network structure.

As before, we generated a large number of random vectors $\boldsymbol{x}_w$ and $\boldsymbol{x}_i$, and plotted the frequency of random matches. With $\| \boldsymbol{x}_w \|_0=k$, we focus on simulations where the non-zero entries in $\boldsymbol{x}_w$ are uniform in $[-1/k, 1/k]$, and the non-zero entries in $\boldsymbol{x}_i$ are uniform in $S*[0, 2/k]$.  We focus on this formulation because of the relationship to common network structures and weight initialization. $\boldsymbol{x}_w$ is a putative weight vector and $\boldsymbol{x}_i$ is an input vector to this layer from the previous layer (we assume unit activations are positive, the result of a ReLU-like non-linearity).  $S$ controls the scale of $\boldsymbol{x}_i$ relative to $\boldsymbol{x}_w$.

Figure~\ref{scalar-sdr} (\textbf{left}) shows the behavior with $k=32$ and $S=1$. We varied the activity of the input vectors $\| \boldsymbol{x}_i \|_0=a$ and the dimensionality of the vectors, $n$.  We set $\theta = E[\boldsymbol{x}_w \cdot \boldsymbol{x}_w] / 2.0$. The chart demonstrates that under these conditions we can achieve robust behavior similar to that of binary vectors. Figure~\ref{scalar-sdr} (\textbf{right}) plots the effect of $S$ on the match probabilities with a fixed $n=1000$.  As this chart shows, the error increases significantly as $S$ increases. Taken together, these results show that the fundamental robustness properties of binary sparse vectors can also hold for sparse scalar vectors, as long as the overall scaling of vectors are in a similar range.

\subsection{Non-uniform Distribution of Vectors}

Eq.~\ref{p1} assumes the ideal case where vectors are chosen with a uniform random distribution. With a non-uniform distribution the error rates will be higher. The more non-uniform the distribution the worse the error rates. For example, if you mostly end up observing 10 inputs, your error rates will be bounded at around $10\%$. Thus, to optimize error rates, it is important to be as close to a uniform distribution as possible. 

\section{Sparse Network Description}
\label{sparsenet}

Here we discuss a particular sparse network implementation that is designed to exploit Eq.~\ref{p1}. This implementation is an extension of our previous work on the HTM Spatial Pooler, a binary sparse coding algorithm that models sparse code generation in the neocortex \cite{Hawkins2011, Cui2017}. Specifically, we formulate a version of the Spatial Pooler that is designed to be a drop-in layer for neural networks trained with back-propagation. Our work is also closely related to previous literature on k-winner take all networks \cite{majani1989k} and fixed sparsity networks \cite{Makhzani2015}.

\begin{figure*}
\vskip 0.2in
\begin{center}
\centerline{\includegraphics[width=\textwidth]{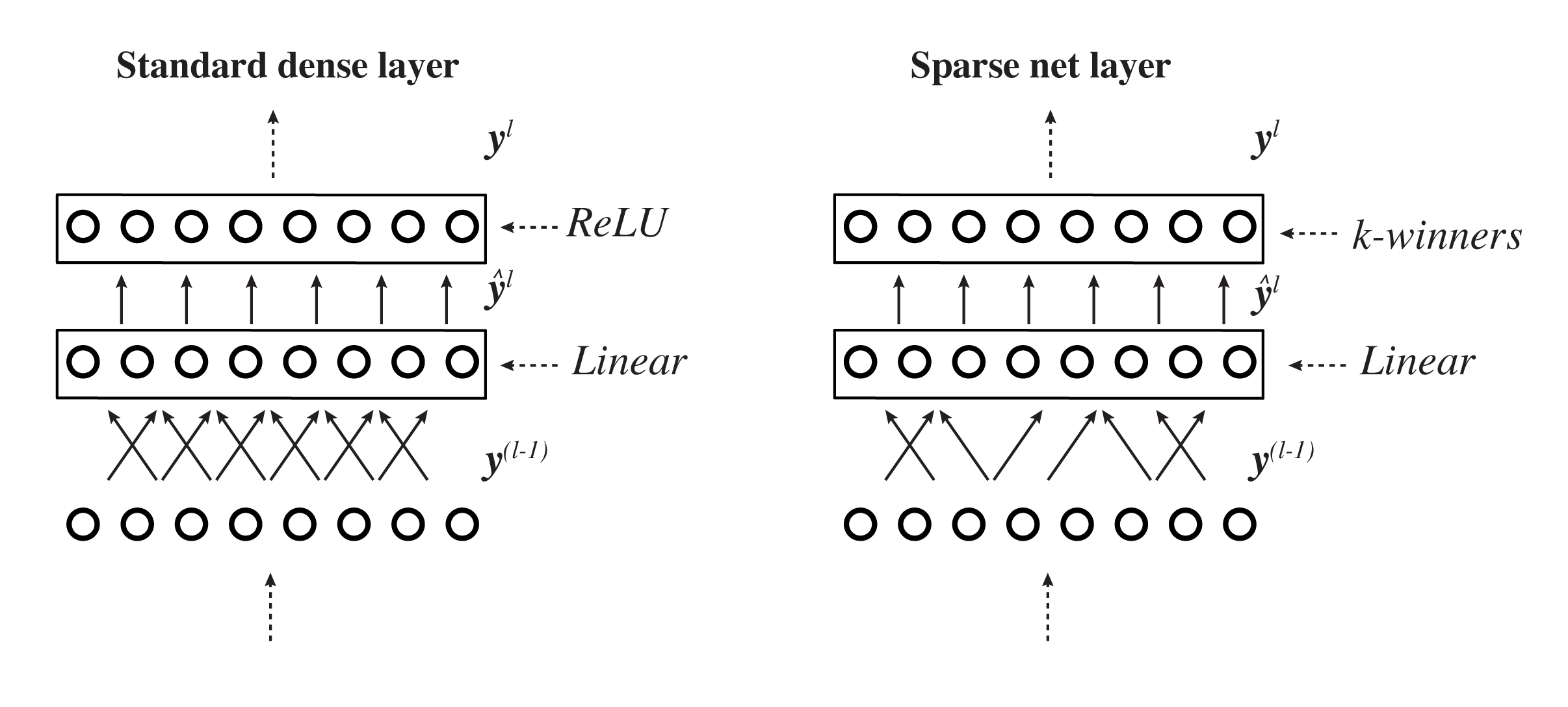}}
\vskip -0.2in
\caption{This figure illustrates the differences between a generic dense network layer (\textbf{left}) and a sparse network layer (\textbf{right}). In the sparse layer, the linear layer subsamples from its input layer (implemented via sparse weights, depicted with fewer arrows). In addition, the ReLU layer is replaced by a k-winners layer.}
\label{net}
\end{center}
%\vskip -0.2in
\end{figure*}

Consider a network with $L$ hidden layers. Let $\boldsymbol{y}^l$ denote the vector of outputs from layer $l$, respectively, with $\boldsymbol{y}^0$ as the input vector. $\boldsymbol{W}^l$ and $\boldsymbol{u}^l$ are the weights and biases for each layer. In a standard neural network the weights $\boldsymbol{W}^l$ are typically dense and initialized using a uniform random distribution. The feed forward outputs are then calculated as follows:

$$\hat{\boldsymbol{y}}^l = \boldsymbol{W}^l \cdot \boldsymbol{y}^{l-1} + \boldsymbol{u}^l$$
$$\boldsymbol{y}^l = f(\hat{\boldsymbol{y}}^l)$$

where $f$ is any activation function, such as $\tanh(\cdot)$ or $\text{ReLU}(\cdot)$. (Figure~\ref{net} left.)

To implement our sparse networks, we make two modifications to this basic formulation (Figure~\ref{net} right.). First, we initialize the weights using a sparse random distribution, such that only a fraction of the weights contain non-zero values. Non-zero weights are initialized using standard Kaiming initialization \cite{He2015}. The rest of the connections are treated as non-existent, i.e. the corresponding weights are zero throughout the life of the network. Second, only the top-k active units within each layer are maintained in $\boldsymbol{y}^l$, and the rest set to zero. This $k$-winners step is non-linear and can be thought of as a substitute for the ReLU function. Instead of a threshold of $0$, the threshold here is adaptive and corresponds to the $k$'th largest activation \cite{Makhzani2013}. 

The layer can be trained using standard gradient descent. Similar to ReLU, the gradient of the layer is calculated as $1$ above the threshold and $0$ elsewhere. During inference we increase $k$ by 50\%, which led to slightly better accuracies. In all our simulations the last layer of each network is a standard linear output layer with log-softmax activation function.

\subsection{Boosting}

One practical issue with the above formulation is that it is possible for a small number of units to initially dominate and then, through learning, become active for a large percentage of patterns (this was also noted in \cite{Makhzani2015, Cui2017}).  Having a small number of active units negatively impacts the available representational volume. It is desirable for every unit to be equally active in order to maximize the robustness of the representation in Eq.~\ref{p1}. To address this we employ a boosting term \cite{Hawkins2011,Cui2017} which favors units that have not been active recently. We compute a running average of each unit's duty cycle (i.e. how frequently it has been one of the top $k$ units):

\begin{equation}
d_i^l (t) = (1-\alpha)d_i^l(t-1) + \alpha \cdot [i \in \text{topIndices}^l]
\end{equation}

A boost coefficient $b_i^l$ is then calculated for each unit based on the target duty cycle and the current average duty cycle:

%TODO: plot this and show in supplemental materials?
\begin{equation}
\label{boost-equation}
b_i^l (t) = e^{\beta( \hat{a}^l - d_i^l(t) )}
\end{equation}

The target duty cycle $\hat{a}^l$ is a constant reflecting the percentage of units that are expected to be active, i.e. $\hat{a}^l = \frac{k}{|\boldsymbol{y}^l|}$. The boost factor, $\beta$, is a positive parameter that controls the strength of boosting.  $\beta = 0$ implies no boosting ($b_i^l = 1$), and higher numbers lead to larger boost coefficients. In \cite{Hawkins2011,Cui2017} we showed that Eq.~\ref{boost-equation} encourages each unit to have equal activation frequency and effectively maximizes the entropy of the layer.

The boost coefficients are used during the k-winners step to select which units remain active for this input. Through boosting, units which have not been active recently have a disproportionately higher impact and are more likely to win, whereas overly active units are de-emphasized. To determine the output of the layer, the non-boosted activity of each winning unit is kept and the remaining units are set to zero. The duty cycle is then updated. The complete pseudo-code description for the $k$-winners layer is described in Algorithm~\ref{alg:kwinners}. In our simulations we used $\beta = 1.0 \text{ or } 1.5$ for all sparse simulations.

\begin{algorithm}[tb]
   \caption{$k$-winners layer}
   \label{alg:kwinners}
\begin{algorithmic}[1]
   \STATE $\hat{\boldsymbol{y}}^l = \boldsymbol{w}^l \cdot \boldsymbol{y}^{(l-1)} + \boldsymbol{u}^l$
   \STATE $b_i^l (t) = e^{\beta( \hat{a}^l - d_i^l(t) )}$
   \STATE $\text{topIndices}^l = topk(\boldsymbol{b}^l \odot \hat{\boldsymbol{y}}^l)$
   \STATE $\boldsymbol{y}^l = 0$
   \STATE $\boldsymbol{y}^l[\text{topIndices}^l] = \hat{\boldsymbol{y}}^l$
   \STATE $d_i^l (t) = (1-\alpha)d_i^l(t-1) + \alpha \cdot [y_i^l(t) \in \text{topIndices}^l]$
\end{algorithmic}
\end{algorithm}

\subsection{Sparse Convolutional Layers}

We can apply the above algorithm to convolutional networks (CNNs) \cite{LeCun1989}. A canonical CNN layer uses a linear convolutional layer containing a number of filters, followed by a max-pooling (downsampling) layer, followed by ReLU. In order to implement sparse CNN layers, the k-winners layer is applied to the output of the max-pooling layer instead of ReLU (just as in our non-convolutional layers). However, since each filter in a CNN shares weights across the image, duty cycles are accumulated per filter. In our simulations dense and sparse CNN nets both have a hidden layer (which is dense or sparse, respectively) after the last convolutional layer, followed by a linear plus softmax layer. We used $5X5$ filters throughout with a stride of $1$. In our tests, the weight sparsity of CNN layers did not impact the results. We suspect this is due to the small size of each kernel and did not use sparse weights for the CNN filters in our experiments.

\section{Results}
\label{results}

\subsection{MNIST}

We first trained our networks on MNIST \cite{LeCun1998}. We trained both dense and sparse implementations. Each network consisted of one or two convolutional layers, followed by a hidden layer, followed by a linear + softmax output layer. Sparse nets consisted of sparse convolutional layers followed by a sparse hidden layer.

Networks were trained using standard stochastic gradient descent to minimize cross entropy loss. We used starting learning rates in the range $0.01-0.04$, and the learning rate was decreased by a factor between 0.5 and 0.9 after each epoch. We also tried batch normalization \cite{Ioffe2015} and found it did not help for MNIST (it did help significantly for Google Speech Commands results - see below). For sparse networks, we used a small mini-batch size (around 4), for the first epoch only, in order to let duty cycle calculations update frequently and settle. Hyperparameters such as the learning rate and network size were chosen using a validation set consisting of $10,000$ randomly chosen training samples. We then report final results on the test set using networks trained on the full training set. 

\textbf{Results Without Noise:} State of the art accuracies on MNIST using convolutional neural networks (without distortions or other training augmentation) are in the range $98.3-99\%$ respectively\footnote{Source: \url{http://yann.lecun.com/exdb/mnist}}. Table~\ref{mnist-table} (left column) lists the classification accuracies for the networks in our experiments. Our accuracies are in the same range, for both sparse and dense networks. Table~\ref{params} lists the key parameters for each of the listed networks (see also the next section for a more in-depth discussion).

\begin{table}[t]
\vskip 0.15in
\begin{center}
\begin{small}
\begin{sc}
\begin{tabular}{lcccr}
\toprule
Network & Test Score & Noise Score \\
\midrule
 dense CNN-1     & 99.14 $\pm$ 0.03 & 74,569 $\pm$ 3,200  \\
 dense CNN-2     & 99.31 $\pm$ 0.06 & 97,040 $\pm$ 2,853  \\
\midrule
 sparse CNN-1    & 98.41 $\pm$ 0.08 & 100,306 $\pm$ 1,735 \\
 sparse CNN-2    & 99.09 $\pm$ 0.05 & 103,764 $\pm$ 1,125 \\
\midrule
Dense CNN-2 SP3  & 99.13 $\pm$ 0.07 & 100,318 $\pm$ 2,762 \\
Sparse CNN-2 D3 & 98.89 $\pm$ 0.13 & 102,328$\pm$ 1,720\\
Sparse CNN-2 W1 & 98.2$\pm$ 0.19 & 100,322$\pm$ 2,082\\
Sparse CNN-2 DSW & 98.92 $\pm$ 0.09 & 70,566 $\pm$ 2,857  \\
\bottomrule
\end{tabular}
\end{sc}
\end{small}
\end{center}
\vskip -0.1in
\caption{MNIST results for dense and sparse architectures. We show classification accuracies and total noise scores (the total number of correct classification for all noise levels). Results are averaged over 10 random seeds, $\pm$ one standard deviation. CNN-1 and CNN-2 indicate one or two convolutional layers, respectively.}
\label{mnist-table}
\end{table}

\begin{figure*}
\vskip 0.2in
\begin{center}
\centerline{\includegraphics[width=\textwidth]{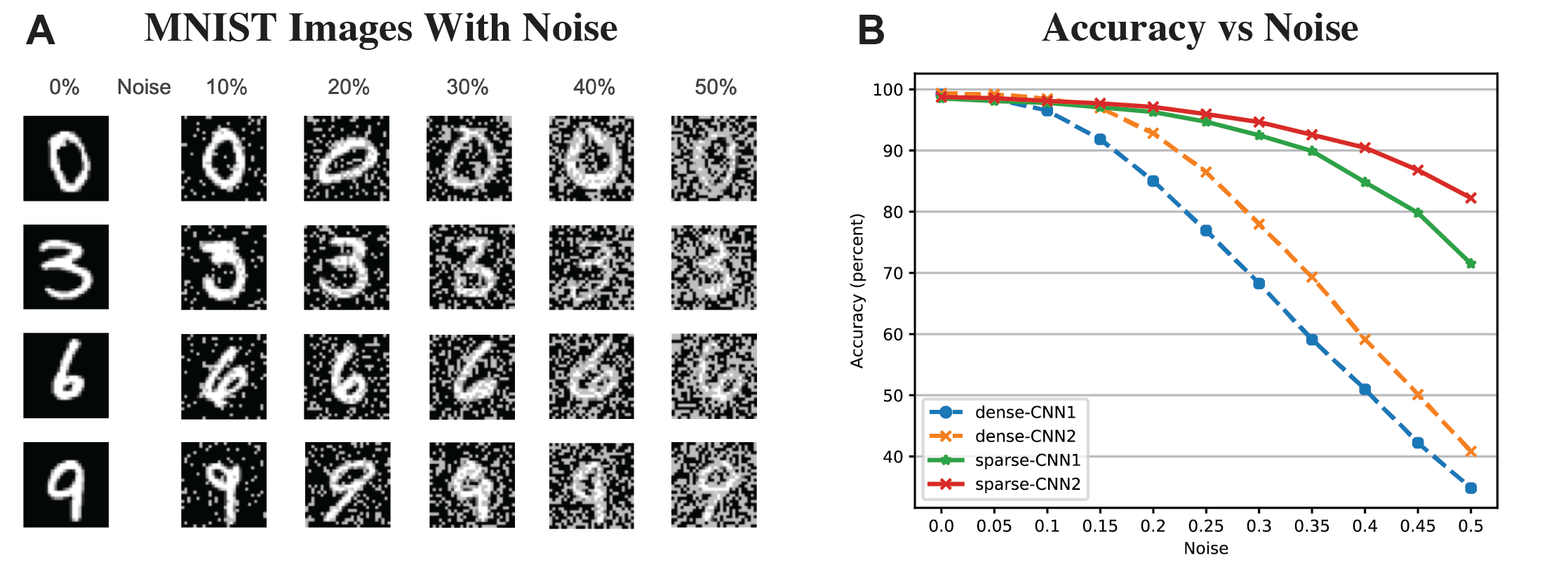}}
\vskip -0.2in
\caption{\textbf{A}. Example MNIST images with varying levels of noise. \textbf{B}. Classification accuracy as a function of noise level.}
\label{mnist-noise}
\end{center}
%\vskip -0.2in
\end{figure*}

\textbf{Results With Noise:} In order to test noise robustness we generated MNIST images with varying levels of additive noise. For each test image we randomly set $\eta\%$ of the pixels to a constant value near white (the constant value was two standard deviations over the mean pixel intensity). Figure~\ref{mnist-noise} (A) shows sample images for different noise levels. We generated $11$ different noise levels with $\eta$ ranging between $0$ and $0.5$ in increments of $0.05$. We also computed an overall \textbf{noise score} which counted the total number of correct classifications across all noise levels.

The right column of Table~\ref{mnist-table} shows the noise scores for each of the architectures. Networks in the top section of the table (Dense CNN-1 and Dense CNN-2) are composed of standard dense convolutional and hidden layers. Networks in the middle section (Sparse CNN-1 and Sparse CNN-2) are composed of sparse convolutional and sparse hidden layers. Networks in the last section contain a mixture of dense and sparse layers. Overall the architectures with sparse layers performed significantly better on the noise score than the fully dense networks. Sparse CNN-2, the two layer completely sparse network, had the best noise score. The two fully dense networks performed substantially worse than the others on noise, even though their test accuracies were comparable. Figure~\ref{mnist-noise} plots the accuracy of fully dense and sparse networks at different noise levels. Note that raw test score was not a predictor of noise robustness, suggesting that focusing on pure test set accuracy alone is not sufficient for gauging performance under adverse conditions.

\textbf{Ablation studies:} In order to judge the relative contributions of sparse layers we ran experiments where we replaced various sparse components with their dense counterparts, i.e. dense CNNs with sparse hidden layers, and vice versa. Dense CNN-2 SP3 contained two dense CNN layers followed by the sparse third layer from Sparse CNN-2. Sparse CNN-2 D3 contained the same CNN layers as Sparse CNN-2 followed by the dense third layer from Dense CNN-2. Sparse CNN-2 W1 was identical to Sparse CNN-2 except that the weight sparsity was 1 (i.e. fully dense weights). Sparse CNN-2 DSW contained a third layer with dense outputs, but with a weight sparsity of $0.3\%$. 

The results of these networks are shown in the bottom third of Table~\ref{mnist-table}. From a noise robustness perspective, most of the variants (except for Sparse CNN-2 DSW) performed well, better than the best pure dense network. This supports the idea that sparsity in many forms may be helpful with robustness. It is interesting to note that the standard deviation of the noise score in these variants was also higher than that of the pure sparse networks. Overall the results with mixed networks were encouraging, and suggest a clear benefit to introducing sparsity at any level.

%TODO \textbf{Impact of dimensionality and k:} How does this relate to the equations?  We would show noise accuracy as a function of n and k. We could do a 3D chart where n goes from 200 to 1000, and k goes from 20 to 200. The y-axis is noise accuracy. 

\textbf{Impact of Dropout:} The above results did not use dropout \cite{Srivastava2014}, which is generally thought to improve robustness. We found that dropout did occasionally improve the robustness of dense networks, but any improvements were modest and the dropout percentage had to be tuned carefully. For sparse nets dropout consistently reduced accuracies. Even with the optimal dropout percentage, the noise scores of dense networks were significantly lower than sparse nets. 
% Figure~\ref{dropout-figure} shows the noise score as a function of dropout percentage for sparse and dense nets. 

%TODO: Hyperparameter sensitivity: How sensitive are the networks? What is the learning speed?

\subsection{Google Speech Commands Dataset}

In order to test sparse nets on a different domain, we applied them to the Google Speech Commands dataset (GSC). This audio dataset was made publicly available in 2017 \cite{speechcommands} and consists of 65,000 one-second long utterances of 30 keywords spoken by thousands of individuals. The dataset contains predefined training, validation, and test sets. 

Reference convolutional nets using ten of the keyword categories (plus artificial "silence" and "unknown" categories created during training augmentation) achieve accuracies in the range  $91-92\%$ \cite{sainath2015convolutional,Tang2017}. In \cite{Tang2017} they demonstrated improved accuracies in the range of $95-96\%$ using residual networks (ResNets \cite{He2015,He2015a}). 

A Kaggle competition using GSC (also limited to 10 categories) took place between November 2017 and early 2018\footnote{\url{https://www.kaggle.com/c/tensorflow-speech-recognition-challenge}}. For our simulations we use the preprocessing code provided by one of the top-10 contestants \cite{Tuguldur} who achieved around $97-97.5\%$ accuracies using variants of ResNet and VGG \cite{Simonyan2014} architectures. Following this implementation, audio samples in our simulations are converted to 32-band Mel spectograms before being fed to the network. During training we augment the data by randomly adjusting the amplitude, speed, and pitch of each training sample, and by randomly shifting and stretching samples in the frequency domain. No data augmentation is performed on the validation or test sets. 

We trained dense and sparse convolutional networks, with hyperparameters chosen based on the validation set. We were able to achieve reasonable accuracies using two convolutional layers, followed by a hidden layer and then a linear + softmax output layer. Our sparse networks had sparse convolutional layers as well as a sparse hidden layer.  Unlike MNIST we found that batch normalization \cite{Ioffe2015} accelerated learning significantly, and we used it for every layer.

Using the above setup we were able to achieve test set accuracies in the range of $96.5-97.2\%$ classifying the ten categories corresponding to the digits "zero" through "nine". Table~\ref{gsc-table} (left column) shows mean accuracy on the test set. Both dense and sparse networks had about the same accuracy. Dropout had a negative effect on the accuracy. Table~\ref{params} lists the key parameters in each network.

\begin{table}[t]
\vskip 0.15in
\begin{center}
\begin{small}
\begin{sc}
\begin{tabular}{lcccr}
\toprule
Network & Test Score & Noise Score \\
\midrule
%ResNet-9    & 97.9$\pm$ 0.2 & 11,421$\pm$ 500\\
Dense CNN-2 (Dr=0.0) & 96.37$\pm$ 0.37 & 8,730$\pm$ 471\\
Dense CNN-2 (Dr=0.5)  & 95.69$\pm$ 0.48 & 7,681$\pm$ 368\\
Sparse CNN-2  & 96.65$\pm$ 0.21 & 11,233$\pm$ 1013\\
Super-sparse CNN-2  & 96.57$\pm$ 0.16 & 10,752$\pm$ 942\\
\bottomrule
\end{tabular}
\end{sc}
\end{small}
\end{center}
\vskip -0.1in
\caption{Classification on Google Speech Commands for a number of architectures. We show test and noise scores, averaged over 10 random seeds, $\pm$ one standard deviation. Dr corresponds to different dropout levels.}
\label{gsc-table}
\end{table}

\textbf{Results With Noise:} As with MNIST, we again created noisy versions of the test set. For each test audio sample $\boldsymbol{A}$ we generated a random white noise sample and blended them together:

$$\boldsymbol{A}^* = (1 - \eta) \boldsymbol{A} + \eta \text{whiteNoise}$$

We generated 11 different noise levels, with $\eta$ ranging from 0 to 0.5 in increments of 0.05. Our overall noise score counted the total number of classifications across all noise levels.

As can be seen in Table~\ref{gsc-table} sparse networks performed significantly better than the best dense network. We included a "Super-Sparse CNN-2" with a significantly sparser hidden layer. The hidden layer for this network had $10\%$ weight sparsity, and a lower output sparsity (Table~\ref{params}). This network had slightly lower noise score, but its score was still significantly higher than that of the dense networks. Overall these results demonstrate that the robustness of sparse networks seen with MNIST can scale to other domains.

\begin{table*}[t]
\vskip 0.15in
\begin{center}
\begin{small}
\begin{sc}
\begin{tabular}{lccccccc}
\toprule
Network & L1 F & L1 sparsity & L2 F & L2 sparsity & L3 N & L3 sparsity & Wt sparsity\\
\midrule
\textbf{MNIST} \\
% Dense-NN & & & & & 500 & 100\% & 100\% \\
Dense CNN-1 & 30 & 100\% & & & 1000 & 100\% & 100\% \\
Dense CNN-2 & 30 & 100\% & 30 & 100\% & 1000 & 100\% & 100\% \\
% Sparse-NN & & & & & 500 & 10\%\\
\\
Sparse CNN-1 & 30 & 9.3\% & & & 150 & 33.3\% & 30\% \\
Sparse CNN-2 & 32 & 8.7\% & 64 & 29.3 \% & 700 & 14.3\% & 30\% \\

\\

Dense CNN-2 SP3 & 30 & 100\% & 30 & 100\% & 700 & 14.3\% & 30\% \\
Sparse CNN-2 D3 & 32 & 8.7\% & 64 & 29.3 \% & 1000 & 100\% & 100\% \\
Sparse CNN-2 W1 & 32 & 8.7\% & 64 & 29.3 \% & 700 & 14.3\% & 100\% \\
Sparse CNN-2 DSW & 32 & 8.7\% & 64 & 29.3 \% & 1000 & 100\% & 30\% \\

\midrule

\textbf{GSC} \\

Dense CNN-2 & 64 & 100\% & 64 & 100\% & 1000 & 100\% & 100\% \\
Sparse CNN-2 & 64 & 9.5\% & 64 & 12.5\% & 1000 & 10\% & 40\% \\
Super Sparse CNN-2 & 64 & 9.5\% & 64 & 12.5\% & 1500 & 6.7\% & 10\% \\

\bottomrule
\end{tabular}
\end{sc}
\end{small}
\end{center}
\vskip -0.1in
\caption{Key parameters for each network. L1F and L2F denote the number of filters at the corresponding CNN layer. L1,2,3 sparsity indicates $k/n$, the percentage of outputs that were enforced to be non-zero. $100\%$ indicates a special case where we defaulted to traditional ReLU activations.  Wt sparsity indicates the percentage of weights that were non-zero. All parameters are available in the source code.}
\label{params}
\end{table*}

% \subsection{Impact of Boosting} 

% To quantify the effects of boosting we computed the entropy of our sparse layers. Using the duty cycle as a proxy for the probability of a unit firing, we compute the total binary entropy of a layer:

% $$H^l = \sum_i - d_i^l \log_2(d_i^l) - (1-d_i^l) \log_2(1-d_i^l)$$

% The maximum entropy would be achieved if each unit had a duty cycle exactly equal to $\sfrac{k}{ |\boldsymbol{L}^i|}$. Table~\ref{entropies} shows entropies for networks trained with varying boost factors. Intuitively the entropy shows how well the sparse layer covers the representational volume, with higher entropies indicating better coverage. Note that maximizing entropy by itself is insufficient as a completely random network would achieve maximum entropy but performance would be at chance. Nevertheless when considering sparse networks entropy is a useful metric to gauge the effectiveness of the representation.

% \begin{table}[t]
% \vskip 0.15in
% \begin{center}
% \begin{small}
% \begin{sc}
% \begin{tabular}{lcccr}
% \toprule
% Network & Dataset & Entropy & Noise Score \\
% \midrule
% Sparse-NN & MNIST  & 96.65  & 11,233\\
% Sparse-CNN & MNIST  & 96.65  & 11,233\\
% Sparse-NN & GSC  & 96.65  & 11,233\\
% Sparse-CNN & GSC  & 96.65  & 11,233\\
% \bottomrule
% \end{tabular}
% \end{sc}
% \end{small}
% \end{center}
% \vskip -0.1in
% \caption{The entropy of each sparse network computed over the test set.}
% \label{entropies}
% \end{table}

\subsection{Computational Considerations}  

In standard networks, the size of each weight matrix is $|\boldsymbol{W}^l| = |\boldsymbol{y}^{l-1}| |\boldsymbol{y}^l|$ and the order of complexity of the feed-forward operation can be approximated by the number of multiplications, $|\boldsymbol{y}^{l-1}|^2 |\boldsymbol{y}^l|$. The computational efficiency of sparse systems is closely related to the fraction of non-zeros. In our sparse hidden layers, both activations and weight values are sparse and the number of non-zero product terms in the forward computation is proportional to $k^{l-1} w^l |\boldsymbol{y}^{l-1}| |\boldsymbol{y}^l|$, where $0 < w^l \leq 1$ is the fraction of non-zero weights. In our convolutional layers, only activations values are sparse and the number of non-zero product terms in the forward computation is proportional to $k^{l-1} * \boldsymbol{K^l*K^l} * |\boldsymbol{y}^l|$, where $\boldsymbol{K^l}$ is the kernel width of each filter. 

As an example, the number of non-zero multiplies between the first two convolutional layers in the GSC Sparse CNN-2 network is $12,544 * 1600 * 6400 = 1.23 X 10^{10}$, about 10.5X smaller than the corresponding dense network. The number of non-zero multiplies between the second convolutional layer and the hidden layer in the same network is $200 * 640,000 * 1000 = 1.28 X 10^{11}$, about 20X smaller than the dense network. For Super Sparse CNN-2, that ratio is 35X as compared to the dense version.

% In our dense networks, the ReLU function also introduces zeros. In order to get a more accurate sense of the number of non-zero multiplications, we used forward function hooks available in PyTorch to count the actual number of non-zeros during the forward operation. Table~\ref{non-zeros} shows the number of non-zero products for MNIST and GSC datasets. 

As can be seen, the number of non-zeros products is significantly smaller in the sparse net implementations. Unfortunately we found that current versions of deep learning frameworks, including PyTorch and Tensorflow do not have adequate support for sparse matrices to exploit these properties, and our implementations ran at the same speed as the corresponding dense networks. We suspect this is due to the fact that highly sparse networks are not sufficiently popular in practice. We hope that studies such as this one will encourage highly optimized sparse implementations. (Note that such optimizations may be non-trivial as the set of $k$-winners changes on every step.) When this becomes feasible our numbers suggest there is a strong possibility for large performance gains and/or improvements in power usage.  It is also worth noting that this reduction in computational complexity does not come at a cost. Rather, our experiments showed that sparse representations can lead to improved accuracies under noisy conditions.

% \begin{table}[t]
% \vskip 0.15in
% \begin{center}
% \begin{small}
% \begin{sc}
% \begin{tabular}{lcccr}
% \toprule
% Network & L1 to L2 nzp & L2 to L3 nzp \\
% \midrule
% % Dense NN     & 96.37 & 8,730\\
% Dense CNN-2    & 95.69 & 7,681 \\
% % Sparse NN    & 96.65 & 11,233 \\
% Sparse CNN-2   & 96.65 & 11,233 \\
% Super-sparse CNN-2   & 96.65 & 11,233 \\
% \bottomrule
% \end{tabular}
% \end{sc}
% \end{small}
% \end{center}
% \vskip -0.1in
% \caption{The approximate number of non-zero products (NZP), for some of our networks for each of the datasets.}
% \label{non-zeros}
% \end{table}

\section{Discussion}
\label{discussion}

In this paper we illustrated benefits of sparse representations. We developed intuitions and theory for the structure of vector matching in the context of binary sparse representations. We then constructed efficient neural network formulations of sparse networks that place internal representations in the sweet spot suggested by the theory. In particular we aim to match sparse activations with sparse weights in relatively high dimensional settings. A boosting rule was used to increase the overall entropy of the internal layers in order to maximize the utilization of the representational space. We showed that this formulation increases the overall robustness of the system to noisy inputs using MNIST and the Google Speech Command Dataset. Both dense and sparse networks showed high accuracies, but the sparse nets were significantly more robust. These results suggest that it is important to look beyond pure test set performance as test accuracy by itself is not a reliable indicator of overall robustness. 

Our work extends the existing literature on sparsity and pruning. A very recent theoretical paper  showed that simple linear sparse networks may be more robust to adversarial attacks \cite{NIPS2018_7308}.  A number of papers have shown that it is possible to effectively introduce sparsity through pruning and retraining \cite{Han2015,Frankle2018,Lee2018a}. The mechanisms introduced here can be seen as complementary to those techniques. Our network enforces sparse weights from the beginning by construction, and sparse weights are learned as part of the training process. In addition, we reduce the overall computational complexity by enforcing sparse activations, which in turn significantly reduces the number of overall non-zero products.  This should produce significant  power savings for optimized hardware implementations.

We demonstrated increased robustness in our networks whereas the papers on pruning typically do not explicitly test robustness. It is possible that such networks are also more robust, though this remains to be tested. Pruning techniques in general are quite orthogonal to ours, and it may be feasible to combine them with the mechanisms discussed here. 

In our work we did not attempt to introduce sparsity into the convolutional filters themselves. \cite{Li2016} have shown it is sometimes possible to remove entire filters from large CNNs suggesting that sparsifying filter weights may also be possible, particularly in networks with larger filters. Introducing sparse convolutions within the context of the techniques in this paper is an area of future exploration. The techniques described here are straightforward to implement and can be extended to other architectures including RNNs. This is yet another promising area for future research.

\subsection{Software}

All code and experiments are available at https://github.com/numenta/htmpapers as open source.

\section*{Acknowledgements}

We thank Jeff Hawkins, Ali Rahimi, and John Berkowitz for helpful discussions and comments.

%\textbf{Do not} include acknowledgements in the initial version of
%the paper submitted for blind review.

\bibliography{ms}

\begin{thebibliography}{33}
\expandafter\ifx\csname natexlab\endcsname\relax\def\natexlab#1{#1}\fi
\expandafter\ifx\csname url\endcsname\relax
  \def\url#1{{\tt #1}}\fi
\expandafter\ifx\csname urlprefix\endcsname\relax\def\urlprefix{URL }\fi

\bibitem[{Ahmad \& Hawkins(2016)}]{Ahmad2016}
Ahmad, S., \& Hawkins, J. (2016).
\newblock {How do neurons operate on sparse distributed representations? A
  mathematical theory of sparsity, neurons and active dendrites}.
\newblock {\em arXiv\/}, (pp. arXiv:1601.00720 [q--bio.NC]).
\newline\urlprefix\url{https://arxiv.org/abs/1601.00720}

\bibitem[{Chen et~al.(2018)Chen, Paiton, \& Olshausen}]{Chen2018b}
Chen, Y., Paiton, D., \& Olshausen, B. (2018).
\newblock {The Sparse Manifold Transform}.
\newblock In S.~Bengio, H.~Wallach, H.~Larochelle, K.~Grauman, N.~Cesa-Bianchi,
  \& R.~Garnett (Eds.) {\em Advances in Neural Information Processing Systems
  31\/}, (pp. 10533--10544). Curran Associates, Inc.

\bibitem[{Cui et~al.(2017)Cui, Ahmad, \& Hawkins}]{Cui2017}
Cui, Y., Ahmad, S., \& Hawkins, J. (2017).
\newblock {The HTM Spatial Pooler – a neocortical algorithm for online sparse
  distributed coding}.
\newblock {\em Frontiers in Computational Neuroscience\/}, {\em 11\/}, 111.
\newline\urlprefix\url{https://www.frontiersin.org/articles/10.3389/fncom.2017.00111/abstract}

\bibitem[{Frankle \& Carbin(2018)}]{Frankle2018}
Frankle, J., \& Carbin, M. (2018).
\newblock {The Lottery Ticket Hypothesis: Finding Sparse, Trainable Neural
  Networks}.
\newline\urlprefix\url{http://arxiv.org/abs/1803.03635}

\bibitem[{Guo et~al.(2018)Guo, Zhang, Zhang, \& Chen}]{NIPS2018_7308}
Guo, Y., Zhang, C., Zhang, C., \& Chen, Y. (2018).
\newblock {Sparse DNNs with Improved Adversarial Robustness}.
\newblock In S.~Bengio, H.~Wallach, H.~Larochelle, K.~Grauman, N.~Cesa-Bianchi,
  \& R.~Garnett (Eds.) {\em Advances in Neural Information Processing Systems
  31\/}, (pp. 240--249). Curran Associates, Inc.

\bibitem[{Han et~al.(2015)Han, Pool, Tran, \& Dally}]{Han2015}
Han, S., Pool, J., Tran, J., \& Dally, W. (2015).
\newblock {Learning both Weights and Connections for Efficient Neural Network}.
\newblock In C.~Cortes, N.~D. Lawrence, D.~D. Lee, M.~Sugiyama, \& R.~Garnett
  (Eds.) {\em Advances in Neural Information Processing Systems 28\/}, (pp.
  1135--1143). Curran Associates, Inc.

\bibitem[{Hawkins et~al.(2011)Hawkins, Ahmad, \& Dubinsky}]{Hawkins2011}
Hawkins, J., Ahmad, S., \& Dubinsky, D. (2011).
\newblock {Cortical Learning Algorithm and Hierarchical Temporal Memory}.
\newline\urlprefix\url{http://numenta.org/resources/HTM{\_}CorticalLearningAlgorithms.pdf}

\bibitem[{He et~al.(2015{\natexlab{a}})He, Zhang, Ren, \& Sun}]{He2015a}
He, K., Zhang, X., Ren, S., \& Sun, J. (2015{\natexlab{a}}).
\newblock {Deep Residual Learning for Image Recognition}.
\newline\urlprefix\url{http://arxiv.org/abs/1512.03385}

\bibitem[{He et~al.(2015{\natexlab{b}})He, Zhang, Ren, \& Sun}]{He2015}
He, K., Zhang, X., Ren, S., \& Sun, J. (2015{\natexlab{b}}).
\newblock {Delving Deep into Rectifiers: Surpassing Human-Level Performance on
  ImageNet Classification}.
\newline\urlprefix\url{http://arxiv.org/abs/1502.01852}

\bibitem[{Ioffe \& Szegedy(2015)}]{Ioffe2015}
Ioffe, S., \& Szegedy, C. (2015).
\newblock {Batch Normalization: Accelerating Deep Network Training by Reducing
  Internal Covariate Shift}.
\newline\urlprefix\url{http://arxiv.org/abs/1502.03167}

\bibitem[{Kanerva(1988)}]{Kanerva1988}
Kanerva, P. (1988).
\newblock {\em {Sparse Distributed Memory}\/}.
\newblock Cambridge, MA: The MIT Press.

\bibitem[{LeCun et~al.(1989)LeCun, Boser, Denker, Henderson, Howard, Hubbard,
  \& Jackel}]{LeCun1989}
LeCun, Y., Boser, B., Denker, J.~S., Henderson, D., Howard, R.~E., Hubbard, W.,
  \& Jackel, L.~D. (1989).
\newblock {Backpropagation Applied to Handwritten Zip Code Recognition}.

\bibitem[{LeCun et~al.(1998)LeCun, Bottou, Bengio, \& Haffner}]{LeCun1998}
LeCun, Y., Bottou, L., Bengio, Y., \& Haffner, P. (1998).
\newblock {Gradient-based learning applied to document recognition}.
\newblock {\em Proceedings of the IEEE\/}.

\bibitem[{Lee et~al.(2008)Lee, Ekanadham, \& Ng}]{Lee2008}
Lee, H., Ekanadham, C., \& Ng, A.~Y. (2008).
\newblock {Sparse deep belief net model for visual area V2}.
\newblock {\em Advances In Neural Information Processing Systems\/}.

\bibitem[{Lee et~al.(2009)Lee, Grosse, Ranganath, \& Ng}]{Lee2009}
Lee, H., Grosse, R., Ranganath, R., \& Ng, A.~Y. (2009).
\newblock {Convolutional deep belief networks for scalable unsupervised
  learning of hierarchical representations}.
\newblock {\em Proceedings of the 26th Annual International Conference on
  Machine Learning - ICML '09\/}, (pp. 1--8).

\bibitem[{Lee et~al.(2018)Lee, Ajanthan, \& Torr}]{Lee2018a}
Lee, N., Ajanthan, T., \& Torr, P. H.~S. (2018).
\newblock {SNIP: Single-shot Network Pruning based on Connection Sensitivity}.
\newline\urlprefix\url{http://arxiv.org/abs/1810.02340}

\bibitem[{Li et~al.(2016)Li, Kadav, Durdanovic, Samet, \& Graf}]{Li2016}
Li, H., Kadav, A., Durdanovic, I., Samet, H., \& Graf, H.~P. (2016).
\newblock {Pruning Filters for Efficient ConvNets}.
\newline\urlprefix\url{http://arxiv.org/abs/1608.08710}

\bibitem[{Majani et~al.(1989)Majani, Erlanson, \& Abu-Mostafa}]{majani1989k}
Majani, E., Erlanson, R., \& Abu-Mostafa, Y.~S. (1989).
\newblock {On the k-winners-take-all network}.
\newblock In {\em Advances in neural information processing systems\/}, (pp.
  634--642).

\bibitem[{Makhzani \& Frey(2013)}]{Makhzani2013}
Makhzani, A., \& Frey, B. (2013).
\newblock {k-Sparse Autoencoders}.
\newline\urlprefix\url{http://arxiv.org/abs/1312.5663}

\bibitem[{Makhzani \& Frey(2015)}]{Makhzani2015}
Makhzani, A., \& Frey, B. (2015).
\newblock {Winner-take-all autoencoders}.
\newblock {\em Advances in Neural Information Processing\/}.
\newline\urlprefix\url{http://papers.nips.cc/paper/5783-winner-take-all-autoencoders}

\bibitem[{Molchanov et~al.(2017)Molchanov, Ashukha, \& Vetrov}]{Molchanov2017}
Molchanov, D., Ashukha, A., \& Vetrov, D. (2017).
\newblock {Variational Dropout Sparsifies Deep Neural Networks}.
\newline\urlprefix\url{http://arxiv.org/abs/1701.05369}

\bibitem[{Nair \& Hinton(2009)}]{Nair2009}
Nair, V., \& Hinton, G.~E. (2009).
\newblock {3D Object Recognition with Deep Belief Nets}.
\newblock In Y.~Bengio, D.~Schuurmans, J.~D. Lafferty, C.~K.~I. Williams, \&
  A.~Culotta (Eds.) {\em Advances in Neural Information Processing Systems
  22\/}, (pp. 1339--1347). Curran Associates, Inc.

\bibitem[{Olshausen \& Field(1997)}]{Olshausen1997}
Olshausen, B.~A., \& Field, D.~J. (1997).
\newblock {Sparse coding with an overcomplete basis set: A strategy employed by
  V1?}
\newblock {\em Vision Research\/}, {\em 37\/}, 3311--3325.

\bibitem[{Rawlinson et~al.(2018)Rawlinson, Ahmed, \& Kowadlo}]{Rawlinson2018}
Rawlinson, D., Ahmed, A., \& Kowadlo, G. (2018).
\newblock {Sparse Unsupervised Capsules Generalize Better}.
\newline\urlprefix\url{http://arxiv.org/abs/1804.06094}

\bibitem[{Rosenfeld et~al.(2018)Rosenfeld, Zemel, \& Tsotsos}]{Rosenfeld2018}
Rosenfeld, A., Zemel, R., \& Tsotsos, J.~K. (2018).
\newblock {The Elephant in the Room}.
\newline\urlprefix\url{http://arxiv.org/abs/1808.03305}

\bibitem[{Sainath \& Parada(2015)}]{sainath2015convolutional}
Sainath, T.~N., \& Parada, C. (2015).
\newblock {Convolutional neural networks for small-footprint keyword spotting}.
\newblock In {\em Sixteenth Annual Conference of the International Speech
  Communication Association\/}.

\bibitem[{Simonyan \& Zisserman(2014)}]{Simonyan2014}
Simonyan, K., \& Zisserman, A. (2014).
\newblock {Very Deep Convolutional Networks for Large-Scale Image Recognition}.
\newline\urlprefix\url{http://arxiv.org/abs/1409.1556}

\bibitem[{Srivastava et~al.(2014)Srivastava, Hinton, Krizhevsky, Sutskever, \&
  Salakhutdinov}]{Srivastava2014}
Srivastava, N., Hinton, G., Krizhevsky, A., Sutskever, I., \& Salakhutdinov, R.
  (2014).
\newblock {Dropout: A Simple Way to Prevent Neural Networks from Overfitting}.
\newblock {\em Journal of Machine Learning Research\/}, {\em 15\/}, 1929--1958.
\newline\urlprefix\url{http://jmlr.org/papers/v15/srivastava14a.html}

\bibitem[{Srivastava et~al.(2013)Srivastava, Masci, Kazerounian, Gomez, \&
  Schmidhuber}]{NIPS2013_5059}
Srivastava, R.~K., Masci, J., Kazerounian, S., Gomez, F., \& Schmidhuber, J.
  (2013).
\newblock {Compete to Compute}.
\newblock In C.~J.~C. Burges, L.~Bottou, M.~Welling, Z.~Ghahramani, \& K.~Q.
  Weinberger (Eds.) {\em Advances in Neural Information Processing Systems
  26\/}, (pp. 2310--2318). Curran Associates, Inc.
\newline\urlprefix\url{http://papers.nips.cc/paper/5059-compete-to-compute.pdf}

\bibitem[{Szegedy et~al.(2013)Szegedy, Zaremba, Sutskever, Bruna, Erhan,
  Goodfellow, \& Fergus}]{Szegedy2013}
Szegedy, C., Zaremba, W., Sutskever, I., Bruna, J., Erhan, D., Goodfellow, I.,
  \& Fergus, R. (2013).
\newblock {Intriguing properties of neural networks}.
\newline\urlprefix\url{http://arxiv.org/abs/1312.6199}

\bibitem[{Tang \& Lin(2017)}]{Tang2017}
Tang, R., \& Lin, J. (2017).
\newblock {Deep Residual Learning for Small-Footprint Keyword Spotting}.
\newline\urlprefix\url{https://arxiv.org/abs/1710.10361}

\bibitem[{Tuguldur(2018)}]{Tuguldur}
Tuguldur, E.-O. (2018).
\newblock pytorch-speech-commands.
\newline\urlprefix\url{https://github.com/tugstugi/pytorch-speech-commands}

\bibitem[{Warden(2017)}]{speechcommands}
Warden, P. (2017).
\newblock {Speech Commands: A public dataset for single-word speech
  recognition.}
\newblock {\em Dataset available from
  http://download.tensorflow.org/data/speech{\_}commands{\_}v0.01.tar.gz\/}.

\end{thebibliography}
\bibliographystyle{apa-good}

% \appendix
% \section{More detailed information}

% Here are some more figures for scalar SDRs and the detailed charts for weight trimming.

\end{document}